# Distilling Knowledge Using Parallel Data for Far-field Speech Recognition

*Jiangyan Yi[1,2], Jianhua Tao[1,2], Zhengqi Wen[1], Bin Liu[1]*

[1]Institute of Automation, Chinese Academy of Sciences, Beijing, China
[2]University of Chinese Academy of Sciences, Beijing, China

{jiangyan.yi, jhtao, zqwen, liubin}@nlpr.ia.ac.cn

## Abstract

In order to improve the performance for far-field speech recognition, this paper proposes to distill knowledge from the close-talking model to the far-field model using parallel data. The close-talking model is called the teacher model. The far-field model is called the student model. The student model is trained to imitate the output distributions of the teacher model. This constraint can be realized by minimizing the Kullback-Leibler (KL) divergence between the output distribution of the student model and the teacher model. Experimental results on AMI corpus show that the best student model achieves up to 4.7% absolute word error rate (WER) reduction when compared with the conventionally-trained baseline models.

**Index Terms**: transfer learning, knowledge distillation, parallel data, deep neural network, far-field speech recognition

## 1. Introduction

In a close-talking setting, automatic speech recognition systems have achieved significant improvement with deep neural network (DNN) based acoustic models [1, 2, 3]. However, far-field speech recognition tasks are still challenging [4], especially when dealing with speech collected from a single distant microphone.

A lot of efforts have been made to improve the performance of far-field speech recognition systems [5, 6, 7]. Many of these approaches use time-synchronize close-talking and far-field parallel data [8, 9, 10].

Some literatures utilize the close-talking data together with the far-field data to train acoustic models for speech recognition. One of the methods is the multi-condition training [10, 11]. This method just uses all the data from different conditions to train acoustic models. The other method is environment-aware training [6, 12]. This approach has been proposed to use close-talking features to help extract environment features as auxiliary information. Other works are proposed to use the enhanced speech to train acoustic models for speech recognition. The dereverberation model is used to estimate the close-talking data given the far-field data [13, 14]. Some researchers [14, 15] train the dereverberation model and the recognition model independently. Others [16, 17, 18, 19] propose a joint training approach between speech enhancement and speech recognition tasks. Moreover, Ravanelli et al. [20] propose a novel network where speech enhancement and speech recognition tasks cooperate with each other.

The above mentioned approaches are able to obtain obvious improvement. However, most of them only use the close-talking data as the training data or the optimized reference. Few of them use the close-talking model to guide the training of the far-field model. More recently, Qian et al. in [10] propose to share knowledge between two hidden layers of the close-talking and the far-field models. This approach achieves promising improvement. However, it only shares knowledge between the hidden layers rather than transfer knowledge between the output layers of the two models.

Therefore, knowledge distillation is proposed to transfer knowledge between the output layers of the close-talking and the far-field models in this paper. The concept of knowledge distillation has been around for a decade [21, 22]. A more general framework is proposed by Hinton et al. [23] to distill knowledge by using high temperature. At a high level, distillation contains training a new model. The new model is trained to mimic the output distribution of a well-trained model.

Similarly, there are several works that use knowledge distillation to compress acoustic models. Li et al. [24] utilize a large DNN model to train a small DNN model. In [25], Chan et al. propose to transfer knowledge from a recurrent neural networks (RNN) model to a small DNN model. Chebotar et al. [26] propose to distill ensembles of acoustic models into a single acoustic model. All of these methods utilize Kullback-Leibler (KL) divergence [27] to minimize the difference of output distributions between the two acoustic models. Previous results show that these methods can compress acoustic models effectively with a little performance loss.

Inspired by the above methods, this paper uses KL divergence to distill knowledge using parallel data to improve the performance for far-field speech recognition. An acoustic model trained with the close-talking data is called a teacher model. An acoustic model trained with the far-field data is called a student model. The student model is trained to imitate the output distribution of the teacher model. The difference between the output distributions of the two models can be minimized by KL divergence. In addition, this paper investigates how the improvement of the student model is influenced by the performance of the teacher models.

The main contributions of this paper are as follows: a) distilling knowledge from the output layer of the close-talking model to the far-field model using KL divergence for far-field speech recognition. b) investigating how the performance of the student model is influenced by different teacher models.

Experimental results on AMI corpus [28] show that the best student model achieves up to 4.7% absolute word error rate (WER) reduction when compared with the conventionally-trained baseline models. The results also show that increases in the accuracy of the teacher model yield similar increases in the performance of the student model.

The rest of this paper is organized as follows. Section 2 describes knowledge distillation using parallel data. Experiments are presented in Section 3. The results are discussed in Section 4. This paper is concluded in Section 5.

## 2. Knowledge distillation using parallel data

In this section, the algorithm of distillation is introduced at first. Then the framework of knowledge distillation for far-field speech recognition is presented in detail.

### 2.1. Distillation

The distillation is to make the teacher model transfer knowledge to the student model. The student model is trained to mimic the output distribution of the teacher model. Thus the student model is forced to be close to the output distribution of the teacher model. This constraint can be realized by minimizing the KL divergence between the output distributions of the two models. Letting $P_c$ denotes the output probabilities of the teacher model, $Q$ denotes the output probabilities of the student model, the difference of the output distributions between the two models is defined as $D_{KL}(P_c||Q)$ which is wished to minimize

$$D_{KL}(P_c||Q) = \sum_i P_c(s_i|x_c) \ln(P_c(s_i|x_c)/Q(s_i|x)) \quad (1)$$

where $i$ denotes the index of senone, $s_i$ denotes the $i$-th senone, $x_c$ is referred as input features of the close-talking speech, $x$ is referred as input features of the far-field speech, $Q(s_i|x)$ denotes the posterior probability of $s_i$ computed from the student model given $x$, $P_c(s_i|x_c)$ denotes the posterior probability of $s_i$ computed from the teacher model given $x_c$. $D_{KL}(P_c||Q)$ can be also defined

$$D_{KL}(P_c||Q) = H(P_c, Q) - H(P_c) \quad (2)$$
$$H(P_c, Q) = \sum_i -P_c(s_i|x_c) \ln Q(s_i|x) \quad (3)$$
$$H(P_c) = \sum_i -P_c(s_i|x_c) \ln P_c(s_i|x_c) \quad (4)$$

Equation (4) is only correlated with the teacher model. So Equation (4) can be neglected. Thus we can define

$$D_{KL}(P_c||Q) \triangleq \sum_i -P_c(s_i|x_c) \ln Q(s_i|x) \quad (5)$$

By Equation (5), we can see that the KL divergence is minimized by minimizing the Cross Entropy (CE) loss function. Thus, the optimization of the distillation can be viewed as the standard CE training criterion. Therefore, the normal backpropagation (BP) algorithm can be directly used to train the student model. The only thing that needs to be changed is that the hard label is replaced with $P_c(s_i|x_c)$. $P_c(s_i|x_c)$ is called soft label.

Equation (5) also indicates that we can still transfer knowledge from the teacher model to the student model, if the loss function or network architecture of the student model is different from the teacher model. It only needs that the output labels of the student model are identical to the teacher model. This approach is a simplified version of the high temperature based distillation proposed by Hinton et al. [23].

### 2.2. Framework of knowledge distillation

The training of the student model is guided by the teacher model using parallel data. The teacher models and the student models are hybrid acoustic models. They have identical output labels which are senones. The framework of knowledge distillation for far-field speech recognition is shown in Fig. 1.

The hard labels $t_{hard}$ are generated from the Gaussian mixture model hidden Markov model (GMM-HMM) model by frame-level. The GMM-HMM model is trained with the close-talking data. The hard labels are one-hot vectors. For example, [0 0 0 1 0 0] denotes the hard labels of one frame.

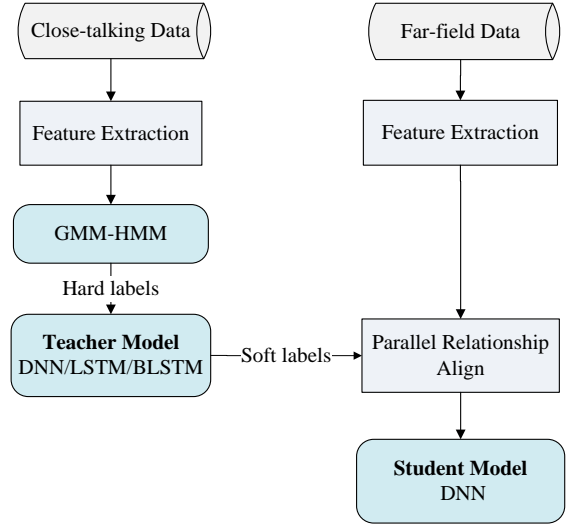

Figure 1: *Framework of knowledge distillation for far-field speech recognition.*

The probability of this frame belonging to label 4 is 1. The probability of this frame belonging to other labels is 0.

The teacher model is trained with the close-talking data $x_c$ and the above hard labels $t_{hard}$. The neural network of the teacher model can be based on DNN [1], long short term memory (LSTM) [29] or bidirectional LSTM (BLSTM). After the training, the parameters of the teacher model are fixed. The teacher model is only used to compute the soft labels.

The soft labels $t_{soft}$ are computed from the teacher model using forward algorithm with the close-talking data $x_c$ by frame-level. The soft labels have much more information about underlying label distribution than the hard labels. For example, [0.01 0.1 0.03 0.79 0 0.07] denotes the soft labels of one frame. The probability of this frame belonging to label 4 is 0.79. The probability of this frame belonging to label 1 is 0.01.

The student model is only DNN based acoustic model. The parallel relationship is used to align the far-field data $x$ and the close-talking soft labels $t_{soft}$. Then the student model is trained using far-field data $x$ with the corresponding soft labels. The training criterion is Equation (5). The parameters of the student model are updated but the parameters of the teacher model aren't changed, when training the student model.

At the decoding stage, only the student model is used to compute posterior probabilities. Then the acoustic likelihood can be computed by combining posterior with prior probabilities. Thus our proposed method doesn't need extra computation cost for decoding.

## 3. Experiments

This section presents experiments to evaluate our proposed approach.

### 3.1. Corpus

Our experiments are conducted on AMI Meeting Corpus [28]. This corpus consists of 100 hours of meeting recordings. The recordings use a range of signals synchronized to a common timeline. There are three types of recordings: IHM, SDM and MDM datasets. IHM is the close-talking data which is collected from individual headset microphones. SDM is the far-field data which is collected from a single distant microphone using 1st microphone array. MDM is the far-field data which is collected from multiple distant microphones using multiple microphones array.

Our experiments only use IHM and SDM datasets. There are three sets for the IHM and SDM datasets respectively: training set (*train*), development set (*dev*) and test set (*eval*). The training set contains 108221 utterances about 80 hours. The development set has 13059 utterances about 10 hours. The test set has 12612 utterances about 10 hours.

### 3.2. Experimental setup

The proposed approach is implemented based on Kaldi speech recognition toolkit [30]. In order to compare our proposed method with the methods in [10], we follow the experimental setup in [10].

The frame length is 25ms and the frame shift is 10ms. The input features of all GMM-HMM models are 39-dim MFCC features. The models have 80K Gaussians. The input features of all neural networks are 40-dimensional log mel-filter bank (FBANK) features plus delta and delta-delta.

The parameters of all the models are updated on the *train* set. The training terminates on the *dev* set with a little improvement. The *dev* set is also used to adjust the hyper parameters and select the models.

The vocabulary is from the AMI dictionary which has 50K words. The language model (LM) is a trigram. The LM is trained using the AMI training transcripts and the Fisher English corpus. The decoding procedure is followed the standard AMI recipe.

### 3.3. Baseline model

We follow the officially released Kaldi recipe to build two GMM-HMM models at first. The Distant-GMM is trained with the far-field data. The Close-GMM is trained with the close-talking data. The Distant-GMM has 4237 senones. The Close-GMM has 4239 senones. Then we use the far-field data from the SDM dataset to train two DNN models. One is called the Distant-DNN which is trained with the hard labels generated from the Distant-GMM using the SDM dataset. The other is called Close-DNN which is trained with the hard labels generated from the Close-GMM using the IHM dataset.

The DNN models have 6 hidden layers with 2048 sigmoid units in each layer. The input layer of the models uses a sliding context window of 11 frames. The models are trained using the stochastic gradient descent (SGD) with mini-batch size of 256. The initial learning rate is set to $1\times 10^{-3}$. The results of the Distant-GMM model and the DNN models on *dev* and *eval* sets of the SDM dataset are listed in Table 1.

Table 1: *WER (%) of three models on the SDM dataset.*

| Model | Hard labels | Dev | Eval |
|---|---|---|---|
| Distant-GMM | - | 64.4 | 69.5 |
| Distant-DNN | SDM | 54.0 | 58.6 |
| Close-DNN | IHM | **50.6** | **55.4** |

From Table 1, we can find that the Close-DNN model outperforms other models on *dev* and *eval* sets obviously. The results show that the use of close-talking hard labels leads to obvious improvement. The reason is that the close-talking hard labels have higher quality than the far-field hard labels. The results are consistent with the conclusions in [10, 31]. Therefore, we select the strongest Close-DNN as the baseline model to compare with our student models.

### 3.4. Close-talking teacher model

There are four teacher models trained using close-talking data from the IHM dataset: DNN, DNN-sMBR, LSTM and BLSTM. The hard labels are generated from the Close-GMM model using the IHM dataset for all the teacher models.

- DNN: This model has the same number of parameters with the baseline Close-DNN model.
- DNN-sMBR: This model is the above DNN model retrained with state-level minimum Bayes risk (sMBR). This model is iterated by 2 epochs.
- LSTM: This model uses a single frame as input. It has 4 stacked LSTM layers with projection, and each layer has 1024 memory cells and 512 output units. The initial learning rate and momentum are set to 0.0001 and 0.9 respectively. The training is carried out by truncated BP through time (BPTT) algorithm.
- BLSTM: This model uses a single frame as input. It has 4 stacked BLSTM layers with projection, and each layer has 512 memory cells and 256 output units. The initial learning rate and momentum are set to 0.0001 and 0.9 respectively. The training is carried out by BPTT algorithm.

The Close-GMM model and the teacher models are evaluated on *dev* and *eval* sets of the IHM dataset. The results of these models are listed in Table 2.

Table 2: *WER (%) of the Close-GMM model and the teacher models on the IHM dataset.*

| Model | Dev | Eval |
|---|---|---|
| Close-GMM | 32.2 | 35.1 |
| DNN | 27.1 | 28.2 |
| DNN-sMBR | 26.0 | 26.1 |
| LSTM | 24.2 | 24.7 |
| BLSTM | **22.5** | **22.8** |

From Table 2, we can find that the BLSTM teacher model achieves the best performance. The LSTM teacher model outperforms all the other DNN teacher models. We use four teacher models to transfer knowledge to the student models in the rest of our experiments.

### 3.5. Far-field student model

All the student models are DNN based acoustic models which have the same number of parameters with the baseline Close-DNN model. There are four student models trained using far-field data from the SDM dataset: S-DNN, S-DNN-sMBR, S-LSTM and S-BLSTM. The teacher model of the S-DNN is the DNN model. The S-DNN-sMBR is trained to mimic the DNN-sMBR teacher model. The teacher model of the S-LSTM is the LSTM model. The S-BLSTM is guided by the DNN-BLSTM teacher model. The soft labels are computed from the teacher models using the IHM dataset respectively.

All the student models are compared with the baseline Close-DNN model. We also compare our proposed method with other methods. DRSL is the method proposed in [14]. Multi-Cond, DRJL-Parallel, DRJL-Front-Back, CFMKS and DRJL+CFMKS are the approaches proposed in [10].

DRSL: training the dereverberation and speech recognition models independently. Multi-Cond: just directly using all the data from close-talking and far-field to train acoustic models. DRJL-Parallel: joint training between the dereverberation and speech recognition models sharing hidden layers. DRJL-Front-Back: joint training between the dereverberation and speech recognition models in front-back structure. CFMKS: sharing knowledge between two hidden layers in the close-talking and the far-field models. All of these models are DNN based. For Multi-Cond, DRJL-Parallel, DRJL-Front-Back, CFMKS, the models are trained using 6 hidden layers with 2048 sigmoid units in each layer. For DRSL and DRJL-Front-Back, both the dereverberation and recognition models are trained using 3 hidden layers with 2048 sigmoid units in each layer respectively. The results of all student models and other models evaluated on the *dev* and *eval* sets of the SDM dataset are listed in Table 3. The WER curves of the student models guided by different teacher models on *eval* set of the SDM dataset are shown in Fig. 2.

From Table 3, we can see that all our student models outperform the baseline model and other models except for the DRJL+CFMKS model. The S-BLSTM student model obtains the best performance among all the models. It achieves 4.7% relative WER reduction on *eval* set when compared with the baseline model, and obtains 2.1% absolute WER reduction on *eval* set over the best model DRJL+CFMKS. The S-DNN student model outperforms the CFMKS model by 0.5% absolute WER reduction, and also outperforms the DRJL-Front-Back model by 0.3% absolute WER reduction on *eval* set.

From Table 3, we also can find that Multi-Cond and DRSL can only obtain a small gain over the baseline. DRJL-Front-Back achieves more improvement than DRSL. These results are consistent with the results in [10].

From Fig. 2, we can see that the student model will achieve better performance, if the teacher model has higher accuracy. The S-BLSTM student model obtains 2.6% absolute WER reduction over the S-DNN student model, when the BLSTM teacher model achieves 5.4% absolute WER reduction over the DNN teacher model on *eval* set.

## 4. Discussion

The above experiments show that our proposed method is effective. Some interesting observations are made as follows.

The best student model outperforms the baseline model and the other conventionally-trained models. There are two main reasons. One is that the teacher model can capture more accurate and better phoneme features from the close-talking data. In contrast, some of the phoneme features from the far-field data are distorted by reverberation and noise. The other is that the soft labels computed from the teacher model contain more information about underlying label distributions when compared with the hard labels. Thus the student model is easier to learn well using more accurate and richer information.

The S-DNN student model outperforms the CFMKS model. The main reason is that the output layers have stronger discriminative ability than the hidden layers. The CFMKS method only shares knowledge between hidden layers. But our proposed method transfers knowledge between output layers.

The S-DNN student model also outperforms the DRJL-Front-Back model. One possible explanation is that the speech may be distorted by the dereverberation model. Nevertheless,

Table 3: *The WER (%) of all student models and other models evaluated on the SDM dataset.*

| Model | Dev | Eval |
|---|---|---|
| Baseline Close-DNN | 50.6 | 55.4 |
| DRSL | 49.8 | 54.8 |
| Multi-Cond | 50.1 | 54.9 |
| DRJL-Parallel | 48.8 | 54.2 |
| DRJL-Front-Back | 47.9 | 53.6 |
| CFMKS | 48.1 | 53.8 |
| DRJL+CFMKS | 47.0 | 52.8 |
| S-DNN | 47.4 | 53.3 |
| S-DNN-sMBR | 46.1 | 52.1 |
| S-LSTM | 45.6 | 51.9 |
| S-BLSTM | **44.5** | **50.7** |

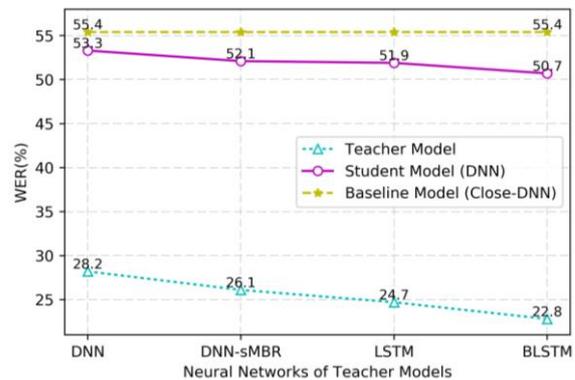

Figure 2: *The WER curves of the student models guided by different teacher models on eval set of the SDM dataset.*

the S-DNN student model only changes its output distribution to imitate the strong teacher model.

In addition, increases in the accuracy of the teacher models yield similar increases in the performance of the student model. If the teacher model has higher accuracy, the student model can train well using more accurate soft labels.

## 5. Conclusions

This paper proposes to distill knowledge from the teacher model to the student model using parallel data to improve the performance of far-field speech recognition tasks. The student model is trained to mimic the output distribution of the teacher model. Thus it can be realized by minimizing the KL divergence between the output distributions of the two models. Experimental results on AMI corpus show that the best student model achieves up to 4.7% absolute WER reduction when compared with the conventionally-trained baseline models. The results also show that increases in the accuracy of the teacher model yield similar increases in the performance of the student model. Moreover, our proposed method doesn't need extra computation cost for decoding. In future work, we plan to use the ensemble teacher model to improve the performance of the student model and apply this approach to other tasks.

## 6. Acknowledgements

This work is supported by the National High-Tech Research and Development Program of China (863 Program) (No.2015AA016305), the National Natural Science Foundation of China (NSFC) (No.61425017, No.61403386).